\documentclass{article}

\usepackage{arxiv}

\usepackage[utf8]{inputenc} 
\usepackage[T1]{fontenc}    
\usepackage{hyperref}       
\usepackage{url}            
\usepackage{booktabs}       
\usepackage{amsfonts}       
\usepackage{nicefrac}       
\usepackage{microtype}      
\usepackage{cleveref}       
\usepackage{lipsum}         
\usepackage{varwidth}
\usepackage{graphicx}
\graphicspath{ {./figures/} }
\usepackage{subcaption} 
\usepackage[square,sort,comma,numbers]{natbib}
\usepackage{doi}

\title{The Bayesian Confidence (BACON) Estimator for Deep Neural Networks}

\author{ \href{https://orcid.org/0000-0000-0000-0000}{\hspace{1mm}Patrick D. Kee}\thanks{DISTRIBUTION STATEMENT A.  Approved for public release.  Distribution is unlimited.  AFRL PA \#: APRS-RYA-2024-05-00011}\\
	Sensors Directorate\\
	Air Force Research Laboratory\\
	Wright-Patterson Air Force Base, Ohio 45433 \\
	\texttt{patrick.kee.1@afrl.af.mil} \\
	\And
	\href{https://orcid.org/0000-0000-0000-0000}{\hspace{1mm}Max J. Brown} \\
	Department of Mathematics\\
	Brigham Young University\\
	Provo, Utah \\
	\AND
	\href{https://orcid.org/0000-0000-0000-0000}{\hspace{1mm}Jonathan C. Rice} \\
	Science, Mathematics, and Engineering Division\\
	UD Sinclair Academy\\
	Dayton, Ohio \\
	\And
	\href{https://orcid.org/0000-0000-0000-0000}{\hspace{1mm}Christian A. Howell} \\
	Department of Mathematics\\
	Brigham Young University\\
	Provo, Utah \\
}

\hypersetup{
pdftitle={The Bayesian Confidence (BACON) Estimator for Deep Neural Networks},
pdfsubject={},
pdfauthor={Patrick D. Kee, Max J. Brown, Jonathan C. Rice, Christian A. Howell},
pdfkeywords={Confidence Estimation, ANN, Bayesian},
}

\begin{document}
\maketitle

\begin{abstract}
    This paper introduces the Bayesian Confidence Estimator (BACON) for deep neural networks. Current practice of interpreting Softmax values in the output layer as probabilities of outcomes is prone to extreme predictions of class probability. In this work we extend Waagen's method of representing the terminal layers with a geometric model, where the probability associated with an output vector is estimated with Bayes' Rule using validation data to provide likelihood and normalization values. This estimator provides superior ECE and ACE calibration error compared to Softmax for ResNet-18 at 85\% network accuracy, and EfficientNet-B0 at 95\% network accuracy, on the CIFAR-10 dataset with an imbalanced test set, except for very high accuracy edge cases. In addition, when using the ACE metric, BACON demonstrated improved calibration error when estimating probabilities for the imbalanced test set when using actual class distribution fractions.
\end{abstract}


\section{Introduction}

The ongoing adoption of machine learning for decision support places a premium in confidence estimation for applications where life, health, and safety are impacted. Reliable confidence estimates are essential for establishing operator trust needed to overcome barriers to adoption. In addition, development of multi-stage machine-to-machine decision support solutions may benefit from use of confidence estimates to weight outputs used as inputs in subsequent stages of processing. The development and characterization of confidence measures has been a subject of research for over three decades. \cite{Bridle:1989}

The Softmax function is widely interpreted as a confidence estimator for deep neural networks. Introduced by Bridle \cite{Bridle:1989} who noted neural network output terms, when optimized through training using a negative log-likelhood loss function, serve in aggregate as probability estimates. He defined the Softmax output activation function such that output terms meet the requirements of a probability, e.g., all terms constrained to zero and one, and all terms sum to unity.

However, the intepretation of Softmax as a probability is not universal \cite{gal:2016}, and deficiencies of Softmax have been noted. Softmax has a tendency to bias predictions towards extremes \cite{Sensoy:2018,Waagen:2020}. These tendencies have been quantified using measures of calibration error, which measure gaps between predicted confidences and true accuracies using binning techniques. These measures include Expected Calibration Error (ECE) using a population weighted measure of the difference between predicted confidence and prediction accuracy; Maximum Confidence Error (MCE), which estimates the maximum deviation between predicted confidence and prediction accuracy \cite{Guo:2017, gong:2021}; and more recently, Adaptive Calibration Error (ACE) uses an adaptive binning approach, creating constant frequency bins which adjust width to capture effects in regions of large populations \cite{Nixon:2019}.

Calibration error has motivated multiple approaches to calibrate confidence estimation. This has included isotonic regression and Platt Scaling \cite{Guo:2017}, Bayesian Binning into Quantiles \cite{Naeini:2015}, and "temperature" scaling, in reference to the superficial resemblance of Softmax to the Boltzmann Probability Distribution (BPD) \cite{Hinton:2014, Guo:2017, posocco:2021}. The superficial resemblance to the BPD resulted in misplaced references to Softmax as the Boltzmann Distribution \cite{Tokic:2011,Tijsma:2016}. \textit{(N.B., unlike Softmax, the BPD has the opposite (negative) sign in the argument, and depends on assumption of a system in equilibrium \cite{Reif:1965}, which is yet unproven for the problem of neural networks.)} 

Confidence estimation for neural networks to date is essentially an empirical approach based on an asserted functional form that has probability characteristics in the aggregate, and sometimes is calibrated in the aggregate. However, these estimates provide limited insight into the dynamics of decision making as individual confidence estimates are based on aggregate characteristics of the training and validation data sets, not on local information that reflects the dynamics of the neural network.

An approach to interpreting dynamics of neural networks was introduced by Waagen, \textit{et al.} \cite{Waagen:2020}. This paper introduced a geometric approach to modeling neural network classification. In this approach, the classification decision is modeled as a projection of the penultimate layer (called the "decision layer") activations onto basis set of output class weight vectors. The position of the decision layer vector in a vector space is determined by the angle associated with the dot product between the decision layer activations and the output layer weight vectors. This geometric picture of the relationship between the final two layers of the neural network provides a representation of the states (classes) and microstates (decision vector orientation) of the neural network that can be the basis of a probability estimation approach based on the internal mechanics of the network. It is then a natural progression of \cite{Waagen:2020} to attempt to estimate the probability of a given class, given the orientation of the decision layer vector. This suggests Bayes' Rule as a possible approach. 

In this paper, we present a novel methodology for confidence estimation by applying a Bayesian approach to the geometric interpretation of classification, by computing the probability that a given geometric configuration corresponds to a particular class. The Bayesian approach also enables the probability estimate to account for imbalance in expected class distributions in the test data set. We present results from an experiment using the CIFAR-10 dataset \cite{PyTorch:2023} used in prior confidence estimation experiments \cite{Guo:2017,Waagen:2020}, and using multiple DNN architectures. Calibration error using multiple error metrics (ECE, MCE, and ACE) are reported, and the merits of the individual metrics are also examined.

The key contribution of this paper to the machine learning body of knowledge is a confidence estimation methodology that is based on the network dynamics instead of an empirical fit, and which also can take into account expected impacts of an imbalanced test set. In addition, we characterize deficiencies in the Softmax confidence estimation approach, as well as identify the deficiencies in the MCE calibration error estimator in the regime of highly accurate classifiers.

\section{Material and Methods}

In this experiment we used the CIFAR-10 dataset \cite{Krizhevsky:2009} obtained from Torchvision \cite{PyTorch:2023}. CIFAR-10 was selected as it has been previously used in similar explorations of confidence estimation errors \cite{Guo:2017,Waagen:2020}. In addition, the small image sizes reduce compute demand, freeing resources for statistical characterization of multliple randomly seeded experiments. 

Our experiment used multiple neural network architectures, ResNet-18 and EfficientNet-B0, both implemented in PyTorch. Sample source code for this experiment is available at our GitHub site at https://github.com/AFRL-RY. Multiple networks were chosen to explore generalizability of results and performance at different classification accuracies. DNNs were implemented using forward hooks to extract and compute node values of the final two layers. Diagnostic tests showed extracted values were pre-activation, not post-activation values. Thus where activation values were required, they were computed post-hoc.

The following process was used to train the neural networks. Networks were trained using GPU-accelerated nodes on three different HP SGI 8600 systems \textit{Koehr, Gaffney, and Mustang} from the DoD High Performance Computing Modernization Program.  For each network architecture, 31 different random seeds were generated to randomly initialize the same number of train/validation splits (in a ratio of 0.8:0.2) of the downloaded CIFAR-10 training data. This resulted in training 31 different sets of weights and biases for each network architecture. Networks were trained using stochastic gradient descent with momentum, minimizing on cross-entropy loss. Best weights and biases based on validation data were stored and updated as training progressed. Training was terminated when completing a set number of epochs.

\begin{figure}[htb]
	\begin{subfigure}[h]{0.4\linewidth}
	\includegraphics[width=\linewidth]{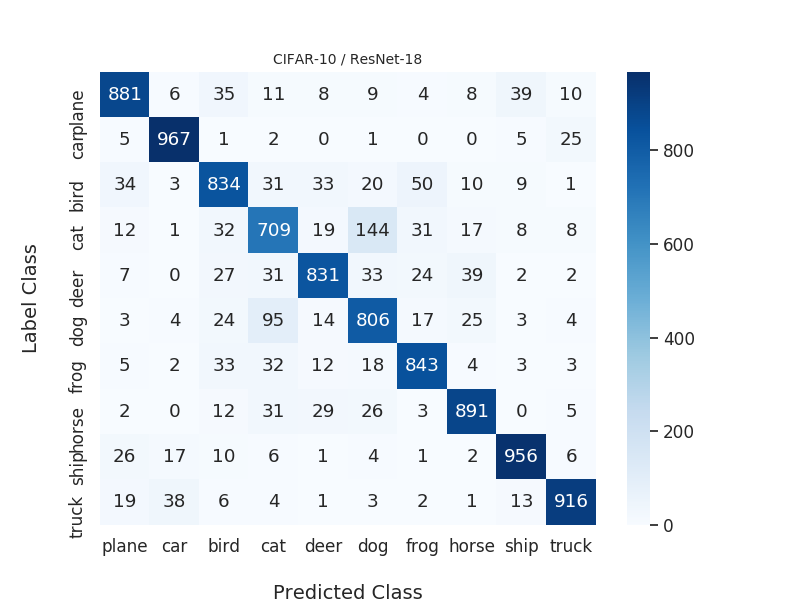}
	\caption{CIFAR 10 / ResNet-18}
	\end{subfigure}
	\hfill
	\begin{subfigure}[h]{0.4\linewidth}
	\includegraphics[width=\linewidth]{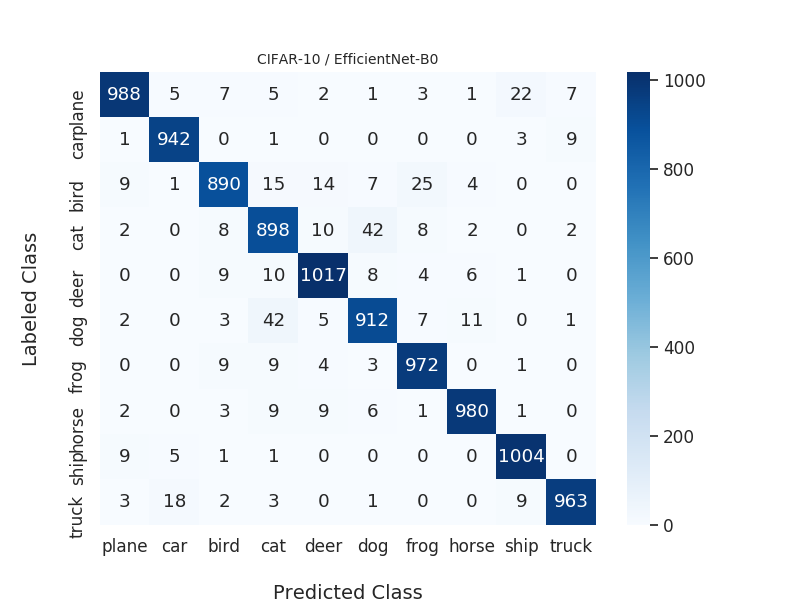}
	\caption{CIFAR-10 / EfficientNet-B0 }
	\end{subfigure}
	\caption{Validation Dataset Confusion Matrices}
	\label{fig:Confusion_Matrices}
	\end{figure}

Best accuracies of networks were $\sim$ 85\% for ResNet-18 and $\sim$ 95\% for EfficientNet-B0. Example confusion matrices are shown in Figure \ref{fig:Confusion_Matrices}. Both ResNet-18 and EfficientNet-B0 matrices are diagonally dominant, EfficientNet-B0 more so than ResNet-18. Of note is the mutual interaction between dog and cat classes. This interaction was exploited in defining our imbalanced test set. 

\section{Experimental}

This experiement used 31 different neural neural networks generated from 31 randomly initilized train/val splits for each architecture to conduct the experiment. For each architecture, thirty networks were used to compute confidences, calibration errors, and z-distribution descriptive statistics. One network for each architecture was held-out for parameter optimization, and results from this seed are not included in reported results.  

At conclusion of training, decision vector angles were computed for the 30 validation datasets on their corresponding networks. These values were used as input values in estimating likelihood and normalization values for Bayesian confidences on the test dataset.

An imbalanced test set was prepared for each randomly initiated network by randomly sampling from the CIFAR-10 test dataset. Test sets were matched by random seed to the corresponding trained network and validation dataset originating from the same random seed. "Dog" and "Cat" classes were shown respectively enhanced and depleted compared to remaining classes to examine impact of using class weights in Bayesian confidence estimates. The distribution of data elements for each class is shown in Table \ref{table:class_distro}.

\begin{table}[htb]
	\caption{CIFAR-10 Imbalanced Test Set Statistics}
	\centering
	\begin{tabular}{lllllllllll}
		\toprule
		\multicolumn{11}{c}{Classes}                   \\
		\cmidrule(r){1-11}
		 & Airplanes & Cars & Birds & Cats & Deer & Dogs & Frogs & Horses & Ships & Trucks \\
		\midrule
		Frequencies & 667 & 667 & 667 & 333 & 667 & 1000 & 667 & 667 & 667 & 667  \\
       \midrule
        Weights & 0.667 & 0.667 & 0.667 & 0.333 & 0.667 & 1.0 & 0.667 & 0.667 & 0.667 & 0.667   \\
		\bottomrule
	\end{tabular}
	\label{table:class_distro}
\end{table}

\section{Theory/Calculation}

\begin{figure}
    \captionsetup{singlelinecheck = false, justification=justified}
	\centering
	\fbox{
	\includegraphics[width=0.5\textwidth]{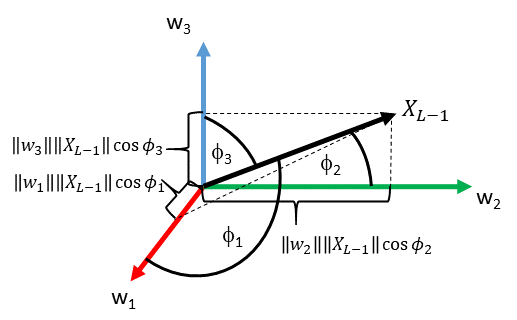}
	}
	\caption{Geometric interpretation of the relationship between decision vector and output layer weight vectors}
	\label{fig:Geometric}
\end{figure}

Our Bayesian approach to estimating confidences is founded on the geometric representation described in \cite{Waagen:2020}, which we will use to define the states and microstates of the system. The geometric representation exploits the use by the feed-forward algorithm of the dot product of present layer activations with next layer weights to compute the node values of the next layer. When the next layer is the output layer, the present layer is referred to as the "decision layer". As weights and biases are fixed at the conclusion of training, all information necessary to classify a particular input are present in the decision layer. The vector of activations of the decision layer can be thought of as a "decision vector", its orientation in space determined by the the angle associated with each output node. Considering a simple case of a network with three output classes, we can visualize the geometric relationship as in Figure \ref{fig:Geometric}. 

Computation of the angles is straightforward, using the dot product relationship between the decision vector and the relevant weight vector. The magnitude of the dot product is related to the product of the magnitude of the factors by the value of a cosine of the angle between the two vectors. Since we can compute the dot product as the sum of the elementwise product of the vectors, as well as the magnitudes of the decision vector and the weight vectors, we can compute an angle associated with each output node with:

\begin{equation}
    \phi_j = cos^{-1} \frac{\left\vert \vec{A*B} \right\vert}{\left\vert \vec{A} \right\vert \left\vert \vec{B} \right\vert}
\end{equation}

In this representation, we interpret the class membership, $j$, of the input data as the state of the system, and the class angle, $\phi_j$, as the microstate of the system. In a Boltzmann-type system, a microstate can only belong to one state.  However, in the present problem, it's possible for a particular microstate to belong to any state.

As we seek to calculate the probablity of state $j$ given the value of angle $\phi$, this suggests using Bayes' Rule to estimate probabilities:

\begin{equation}
    P(j|\phi_j) = \frac{P(\phi_j|j)w_j}{\sum_i w_i P(\phi_i|i)}
\end{equation}

where \begin{math} w_i \end{math}, and  \begin{math} w_j \end{math}, are respectively the class weights for class \begin{math} i \end{math} and class \begin{math} j \end{math}, and the denominator is summed over all class weights.

\begin{figure}
	\centering
	\fbox{
	\includegraphics[width=0.5\textwidth]{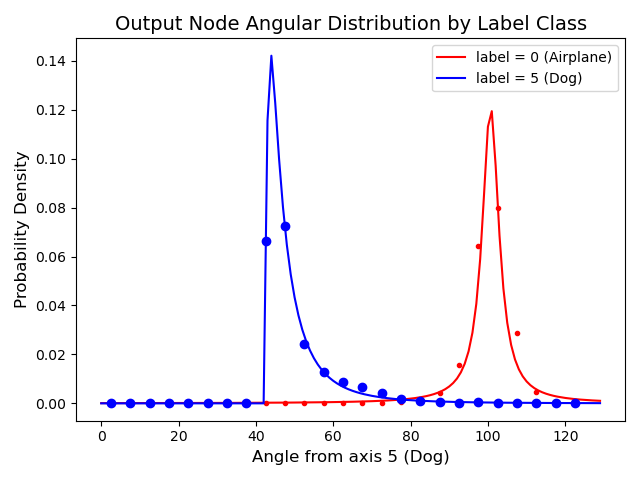}
	}
	\caption{Output node angle distributions by label class for the same output node.}
	\label{fig:Output_Angle_Distribution}
\end{figure}

The output node angle probabilities in Bayes' Rule, \begin{math} P(\phi_i|i) \end{math} are estimated from the validation data set. Angles are computed for each input record, with each combination of output node and labeled classes histogrammed and modeled using an appropriate probability density function (PDF) (see Figure \ref{fig:Output_Angle_Distribution}). For this case, for the output node for "Dog" we modeled a Log Normal distribution for the true label class (Dog), and a Cauchy distribution for a false label class (Airplane). Here it is noted that the angle distribution corresponding to the true label class (Dog) is more abundant at smaller angles than the false label class (Airplane). This is as expected. Referencing Figure \ref{fig:Geometric}, smallest angles correspond to largest output node values, and will tend to correspond to predicted class, neglecting occasional bias term effects.

\begin{figure}
	\centering
	\fbox{
	\includegraphics[width=0.5\textwidth]{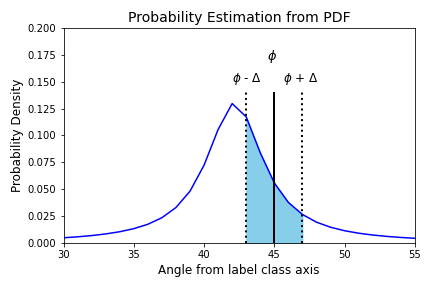}
	}
	\caption{Probability estimation from probability density function}
	\label{fig:Probability_PDF}
\end{figure}

Input probabilities for Bayes' Rule are estimated for angles of interest by integrating the PDF in the neighborhood of the angle, \begin{math} \phi_i \pm \delta \end{math}, with the value of \begin{math} \delta \end{math} optimized to minimize ECE on a hold-out dataset. In practice, the integration is calculated by differencing the the cumulative probability distribution of the PDF over the range of the chosen neighborhood (see Figure \ref{fig:Probability_PDF}).

Bayesian probabilities are compared in this experiment to both Softmax and temperature scaled Softmax (T-scaled Softmax). The general expression is:

\begin{equation}
	Softmax = \frac{exp(\beta x_j)}{\sum_i exp(\beta x_i)}
\end{equation}

where $\beta$ = 1 for unscaled Softmax, while $\beta$ = 1/T is an adjustable parameter for T-scaled Softmax. Values of $\beta$ are chosen to minimize negative loss-likelihood on a holdout dataset \cite{Guo:2017}.

Calibration metrics used in this experiment include the ECE, MCE, and ACE as previously reported by \cite{Guo:2017,Nixon:2019}. These metrics are derived from the reliability diagram described in \cite{DeGroot:1982,DeGroot:1983}. Calibration metrics are based on deviations of binned quantities from an expected value. Confidence estimates for predicted classes are binned, with number of bins, $M$, is equal to $K-1$, where $K$ is the number of classes. The binned average predicted confidence are compared to the average predicted accuracy. The bin values for accuracy and confidence are defined by:

\begin{equation}
    acc = \frac{1}{\left\vert B_m \right\vert} \sum{u \in B_m}^M \mathbf{1} \left ( \hat{y}_i - y_i \right )
\end{equation}

\begin{equation}
    conf = \frac{1}{\left\vert B_m \right\vert} \sum_{i \in B_m}^M \hat{p}_i
\end{equation}

where m is the \begin{math} m_{th} \end{math} bin number, \begin{math}\left\vert B_m \right\vert \end{math} is the frequency of the \begin{math} m_{th} \end{math} bin, \begin{math} M \end{math} is the total number of bins, \begin{math} \hat{y}_i \end{math} is the predicted label for the \begin{math} i_{th} \end{math} point, \begin{math} y_i \end{math} is the true label for the \begin{math} i_{th} \end{math} point, and \begin{math} \hat{p}_i \end{math} is the \begin{math} i^{th} \end{math} confidence estimate in bin \begin{math} B_m \end{math}.

\begin{figure}
	\centering
	\fbox{
	\includegraphics[width=0.5\textwidth]{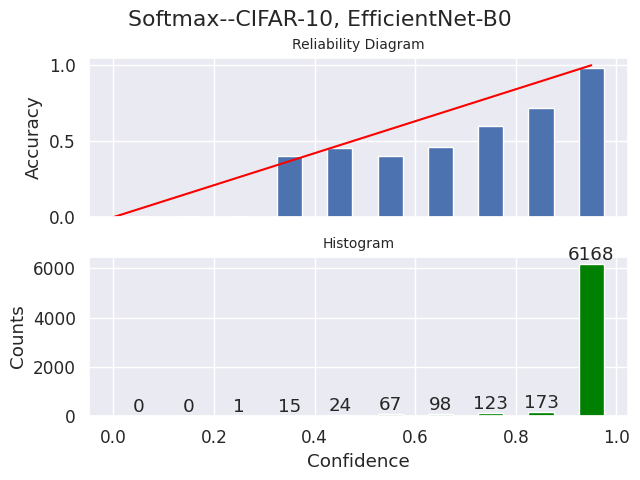}
	}
	\caption{Softmax Reliability Diagram for EfficientNet-B0 at 95\% accuracy}
	\label{fig:Reliability_Diagram}
\end{figure}

In the upper half of Figure \ref{fig:Reliability_Diagram}, the vertical position of the blue bars indicates the bin accuracy. For a well-calibrated confidence estimator, the height of the blue bar should be very close to the red line indicating where accuracy equals confidence.

Our first metric, ECE, combines the results of the reliability diagram with the histogram in the the bottom half of Figure \ref{fig:Reliability_Diagram}. Conceptually, it can be thought of as the sum over the elementwise product of the histogram bin frequencies with the gaps between bin confidence and bin accuracies. It is defined in \citep{Sensoy:2018, Naeini:2015} as:

\begin{equation}
    ECE = \sum_{m=1}^M \frac{\left\vert B_m \right\vert}{n} \left\vert acc \left ( B_m \right ) - conf \left ( B_m \right ) \right\vert
\end{equation}

The second metric, MCE, is defined as the largest deviation between accuracy and confidence for any bin \citep{Sensoy:2018, Naeini:2015}:

\begin{equation}
    MCE = \max_{m \in \left \lbrace 1,...M \right \rbrace} \left \vert acc(B_m) - conf(B_m) \right \vert
\end{equation}

The ACE metric was also used in this effort due to the known problem with ECE at higher neural network accuracies. Due to ECE's use of fixed bins, predictions tend to cluster in the rightmost bins. To add more resolution in high accuracy conditions, adaptive binning has been tried in characterizing confidence calibration \cite{Nixon:2019,posocco:2021}. Nixon, et al, introduced the ACE metric which is similar to ECE, but which uses bins of constant frequency, and includes all class confidence values above a minimum threshold (e.g., threshold = 0.001), not just confidences for the predicted class. ACE is defined as \cite{Nixon:2019}:

\begin{equation}
    ACE = \frac{1}{KR} \sum_{k=1}^K \sum_{r=1}^R \left\vert acc \left ( r,k \right ) - conf \left ( r,k \right ) \right\vert
\end{equation}

where r is the index adaptive calibration range, k is the index of label class, R is the number of calibration ranges, and K is the number of label classes. In this experiment we used the L1 norm when computing ACE.

\section{Results}

We present results for BACON (both unweighted and weighted) and Softmax (unscaled and temperature scaled) for \textit{ResNet-18} and \textit{EfficientNet-B0}, with calibration errors estimated using ECE, ACE, and MCE.  ECE and ACE will be examined first as an entire dataset, and next on a class-by-class basis.

\subsection{Entire Dataset Metrics}

\subsubsection{Expected Calibration Error}

\begin{figure}[htb]
	\begin{subfigure}[h]{0.4\linewidth}
		\includegraphics[width=\linewidth]{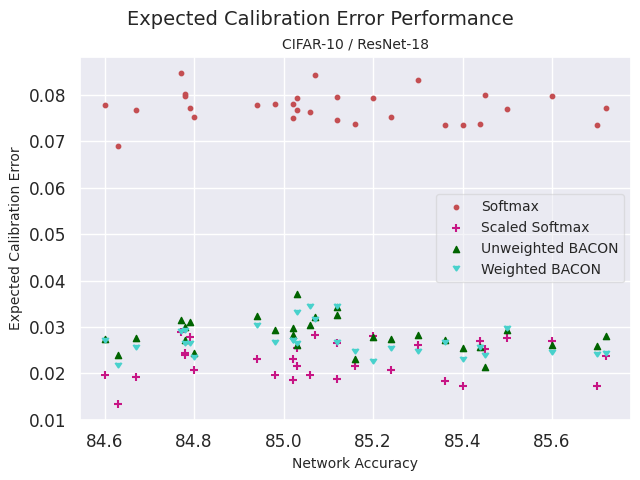}
		\caption{ResNet-18 Individual Seed ECE Results}
	\end{subfigure}
	\hfill
	\begin{subfigure}[h]{0.4\linewidth}
		\includegraphics[width=\linewidth]{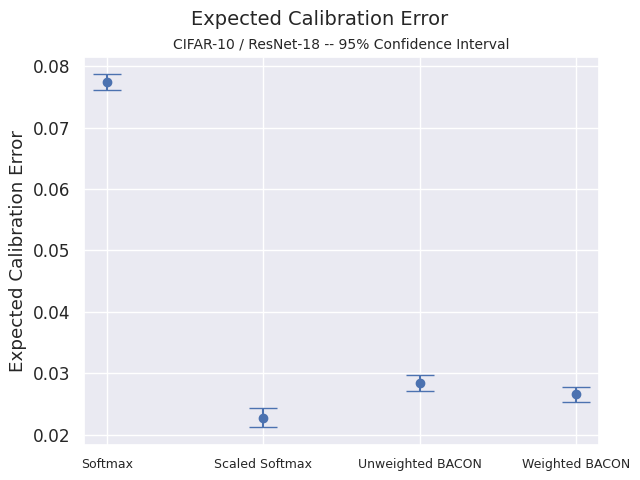}
		\caption{Resnet-18 ECE Confidence Intervals}
	\end{subfigure}
	\caption{ResNet-18 ECE Results}
	\label{fig:ECE_Results_ResNet-18}
\end{figure}

\begin{figure}[htb]
	\begin{subfigure}[h]{0.4\linewidth}
		\includegraphics[width=\linewidth]{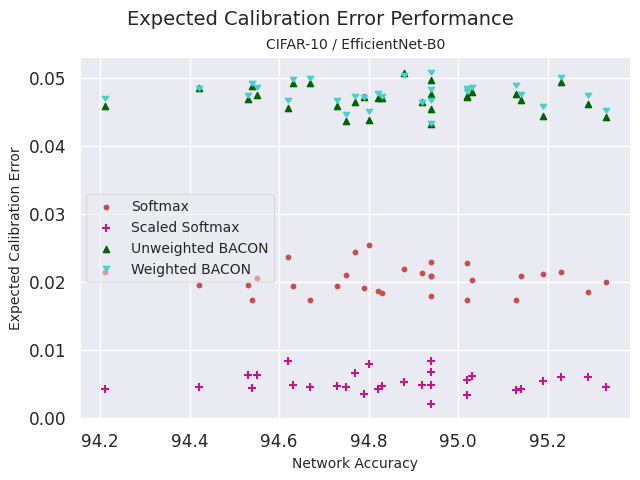}
		\caption{EfficientNet-B0 Individual Seed ECE Results}
	\end{subfigure}
	\hfill
	\begin{subfigure}[h]{0.4\linewidth}
		\includegraphics[width=\linewidth]{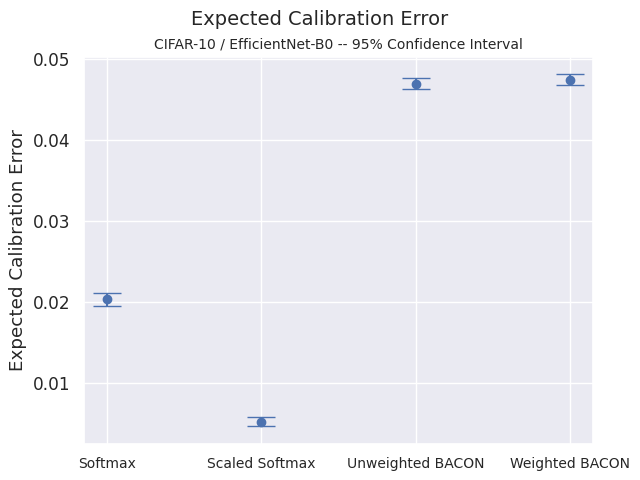}
		\caption{EfficientNet-B0 ECE Confidence Intervals}
	\end{subfigure}
	\caption{EfficientNet-B0 ECE Results}
	\label{fig:ECE_Results_EfficientNet-B0}
\end{figure}

ECE metrics are computed over the entire dataset. Results for all classes are summarized in Figures \ref{fig:ECE_Results_ResNet-18} and \ref{fig:ECE_Results_EfficientNet-B0} and Table \ref{table:ECE_Results}. Results are mixed. For ECE and ResNet-18 at 85\% accuracy, unweighted and weighted BACON both outperform unweighted Softmax, however, T-scaled Softmax performs best of all confidence estimators. For ECE and EfficientNet-B0 at 95\% accuracy, both unweighted and weighted BACON perform worse than both scaled and unscaled versions of Softmax. Also, weighting BACON does not result in a significant difference for either ResNet-18 or EfficientNet-B0.

\begin{table}[htb]
	\caption{ECE Results}
	\centering
	\begin{tabular}{lccccc}
		\toprule
		\multicolumn{6}{c}{ECE $\pm$ 2$\sigma$}                   \\
		\cmidrule(r){1-6}
		& Nominal &  & T-Scaled & Unweighted & Weighted \\
		Network & Accuracy (\%) & Softmax & Softmax & BACON & BACON \\ 
		\midrule
		ResNet-18 & 85 & 0.0774 $\pm$ 0.001 & 0.023 $\pm$ 0.001 & 0.028 $\pm$ 0.001 & 0.027 $\pm$ 0.001 \\
        \midrule
        EfficientNet-B0 & 95 & 0.0204 $\pm$ 0.0008 & 0.0053 $\pm$ 0.0005 & 0.0469 $\pm$ 0.0007 & 0.0474 $\pm$ 0.0007  \\
		\bottomrule
	\end{tabular}
	\label{table:ECE_Results}
\end{table}

\begin{figure}[htb]
\begin{subfigure}[h]{0.4\linewidth}
\includegraphics[width=\linewidth]{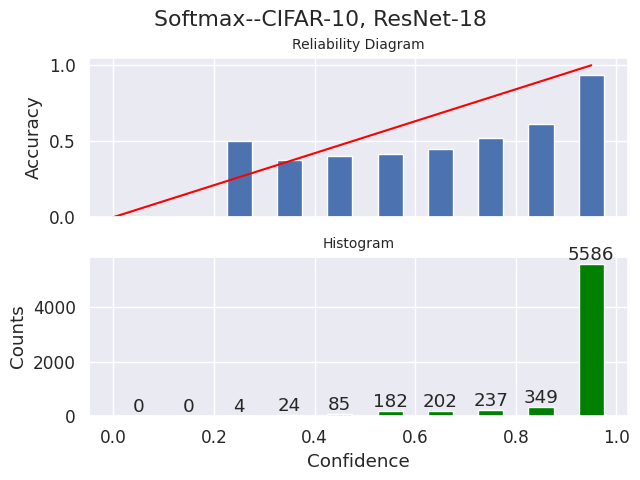}
\caption{Softmax Fixed Bin Reliability Diagram}
\end{subfigure}
\hfill
\begin{subfigure}[h]{0.4\linewidth}
\includegraphics[width=\linewidth]{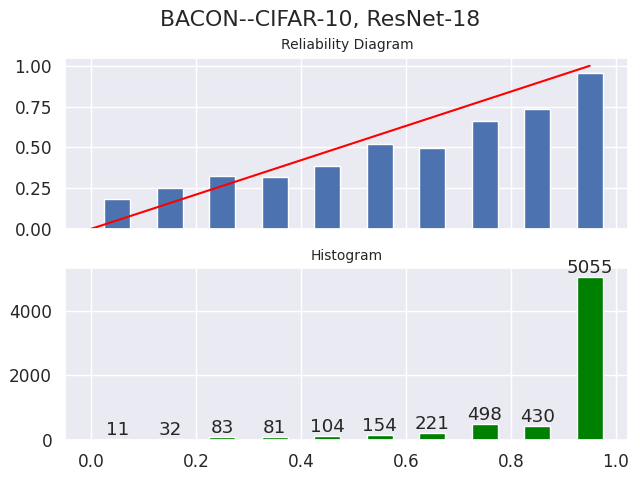}
\caption{Weighted BACON Fixed Bin Reliability Diagram}
\end{subfigure}
\caption{ECE Reliability Diagrams for Softmax and Weighted BACON for ResNet-18}
\label{fig:BACON_Softmax_Reliability_Diagrams_ResNet}
\end{figure}

\begin{figure}[htb]
\begin{subfigure}[h]{0.4\linewidth}
\includegraphics[width=\linewidth]{Softmax_CIFAR-10_EfficientNet-B0_95Reliability_Diagram.png}
\caption{Softmax Fixed Bin Reliability Diagram}
\end{subfigure}
\hfill
\begin{subfigure}[h]{0.4\linewidth}
\includegraphics[width=\linewidth]{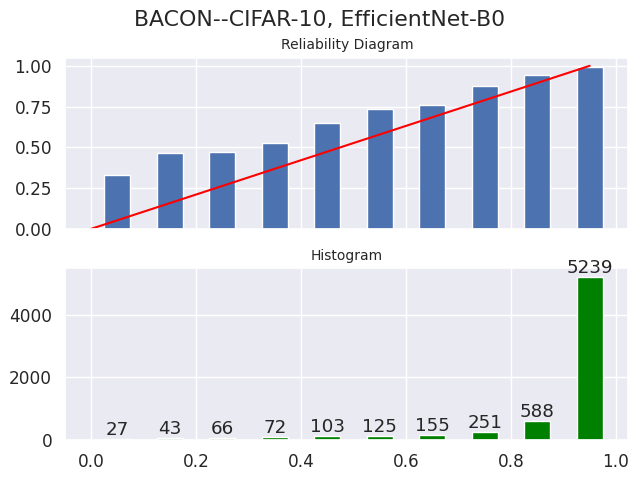}
\caption{Weighted BACON Fixed Bin Reliability Diagram}
\end{subfigure}
\caption{ECE Reliability Diagrams for Softmax and Weighted BACON for EfficientNet-B0}
\label{fig:BACON_Softmax_Reliability_Diagrams_EffNet}
\end{figure}

We next examine edge cases at low and high class accuracies, choosing randomly selected seed 7831 for ResNet-18 and seed 8788 for EfficientNet-B0. The ECE reliability diagrams for Resnet-18 at 85\% network accuracy are shown in Figures \ref{fig:BACON_Softmax_Reliability_Diagrams_ResNet}(a,b). The reliability diagram for BACON appears to have accuracies that track well with predicted confidence, while the plot for Softmax shows overly confident predicted confidences compared to accuracies. Comparison of the two histograms shows that Softmax tends towards higher confidence values compared to BACON. Here, the numerical evaluation of the ECE metric reflects Softmax's larger accuracy/confidence gap for confidences between 0.5 and 1.0, as well as the larger bin counts in this region compared to BACON. Turning our attention to the ECE reliability diagrams for EfficientNet-B0 at 95\% network accuracy in Figures  \ref{fig:BACON_Softmax_Reliability_Diagrams_EffNet}(a,b), we see a different result. Softmax continues to show overprediction in the range of 0.5 through 0.9, while BACON shows underprediction for confidence values less than 0.7. Also notable are the histograms showing Softmax's greater tendency towards high confidence predictions compared to BACON. However, Softmax benefits from beneficial population weighting when ECE is numerically evaluated. Here, the systematic underpredictions for BACON penalize its ECE metric more than the Softmax overpredictions affect its ECE metric.

\subsubsection{Adaptive Calibration Error}

\begin{figure}[htb]
	\begin{subfigure}[h]{0.4\linewidth}
		\includegraphics[width=\linewidth]{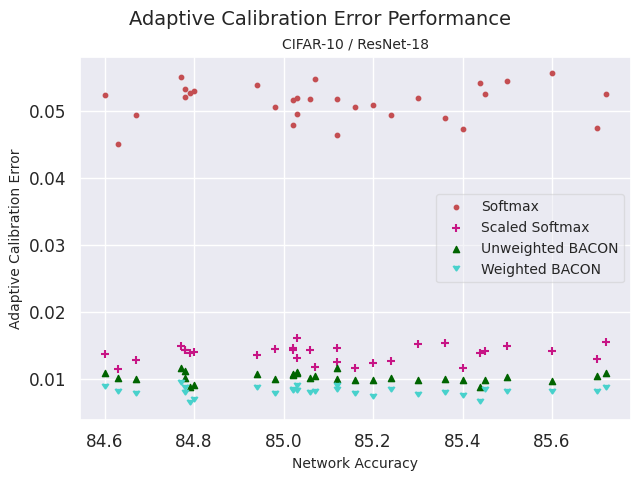}
		\caption{ResNet-18 Individual Seed ACE Results}
	\end{subfigure}
	\hfill
	\begin{subfigure}[h]{0.4\linewidth}
		\includegraphics[width=\linewidth]{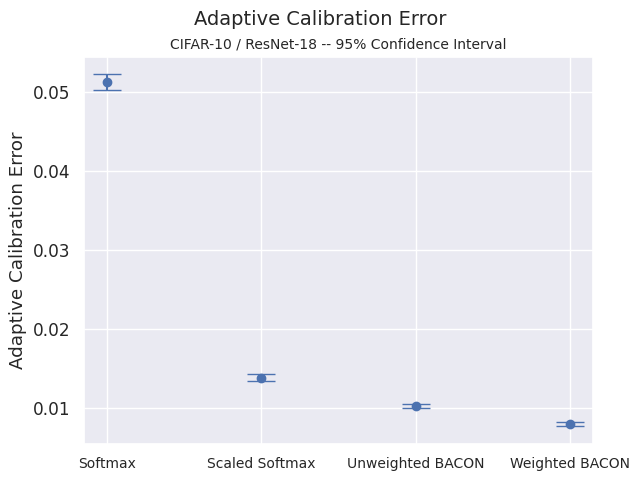}
		\caption{ResNet-18 ACE Confidence Intervals}
	\end{subfigure}
	\caption{ResNet-18 ACE Results}
	\label{fig:ACE_Results_ResNet-18}
\end{figure}

\begin{figure}[htb]
	\begin{subfigure}[h]{0.4\linewidth}
		\includegraphics[width=\linewidth]{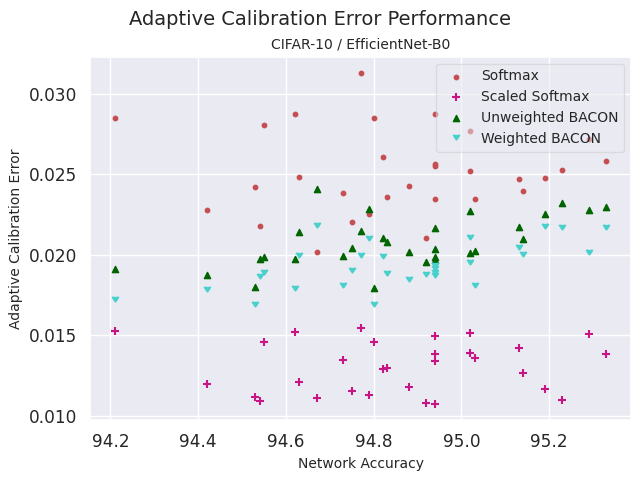}
		\caption{EfficientNet-B0 Individual Seed ACE Results}
	\end{subfigure}
	\hfill
	\begin{subfigure}[h]{0.4\linewidth}
		\includegraphics[width=\linewidth]{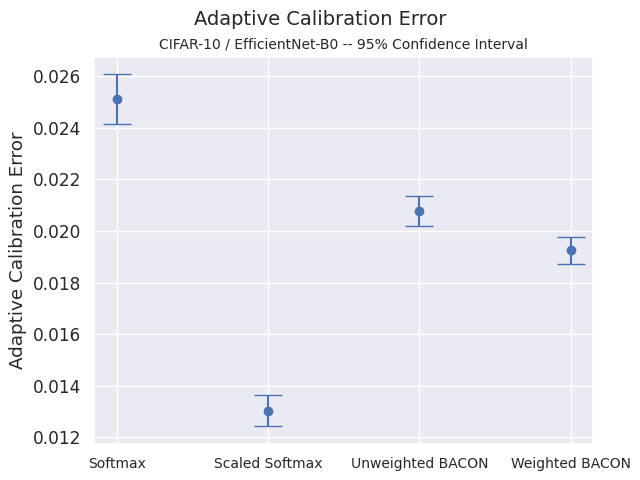}
		\caption{EfficientNet-B0 ACE Confidence Intervals}
	\end{subfigure}
	\caption{EfficientNet-B0 ACE Results}
	\label{fig:ACE_Results_EfficientNet-B0}
\end{figure}

Adaptive calibration error results were computed over the entire dataset (all classes included) and are summarized in Figures \ref{fig:ACE_Results_ResNet-18} and \ref{fig:ACE_Results_EfficientNet-B0} and Table \ref{table:ACE_Results}. Results are mixed. For ECE and ResNet-18 at 85\% accuracy, unweighted and weighted BACON both outperform unweighted Softmax, however, T-scaled Softmax performs best of all confidence estimators. For ECE and EfficientNet-B0 at 95\% accuracy, both unweighted and weighted BACON perform worse than both scaled and unscaled versions of Softmax. For ACE and ResNet-18 at 85\% accuracy, both weighted and unweighted BACON estimators outperform both scaled and unscaled Softmax estimators, in this case, slightly outperforming T-scaled Softmax, with weighted BACON performing best. However, for ACE and EfficientNet-B0 at 95\% accuracy, performance degrades with both weighted and unweighted BACON estimators underperforming both scaled and unscaled Softmax estimators, with T-scaled Softmax performing best. Weighting BACON estimates provides measurable improvement in calibration error compared to unweighted BACON for both ResNet-18 and EfficientNet-B0.

\begin{table}[htb]
	\caption{ACE Results}
	\centering
	\begin{tabular}{lccccc}
		\toprule
		\multicolumn{6}{c}{ACE $\pm$ 2$\sigma$}                   \\
		\cmidrule(r){1-6}
		& Nominal &  & T-Scaled & Unweighted & Weighted \\
		Network & Accuracy (\%) & Softmax & Softmax & BACON & BACON \\ 
		\midrule
		ResNet-18 & 85 & 0.051 $\pm$ 0.001 & 0.0138 $\pm$ 0.005 & 0.0103 $\pm$ 0.0003 & 0.0080 $\pm$ 0.0003 \\
        \midrule
        EfficientNet-B0 & 95 & 0.025 $\pm$ 0.001 & 0.0130 $\pm$ 0.0006 & 0.0208 $\pm$ 0.0006 & 0.0193 $\pm$ 0.0005  \\
		\bottomrule
	\end{tabular}
	\label{table:ACE_Results}
\end{table}

\begin{figure}[htb]
\begin{subfigure}[h]{0.4\linewidth}
\includegraphics[width=\linewidth]{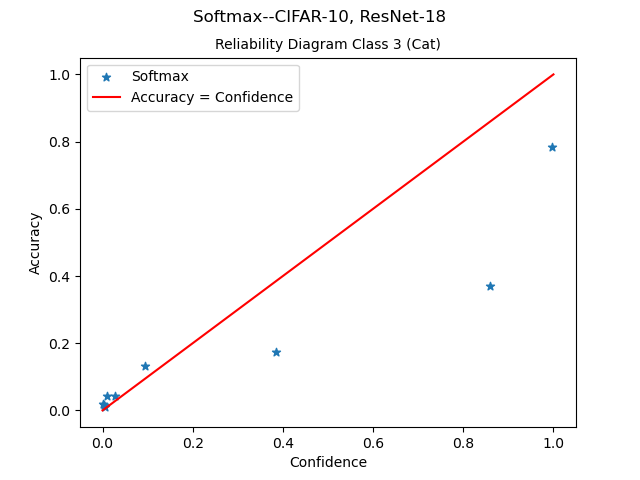}
\caption{Softmax ResNet-18 ACE Reliability Diagram}
\end{subfigure}
\hfill
\begin{subfigure}[h]{0.4\linewidth}
\includegraphics[width=\linewidth]{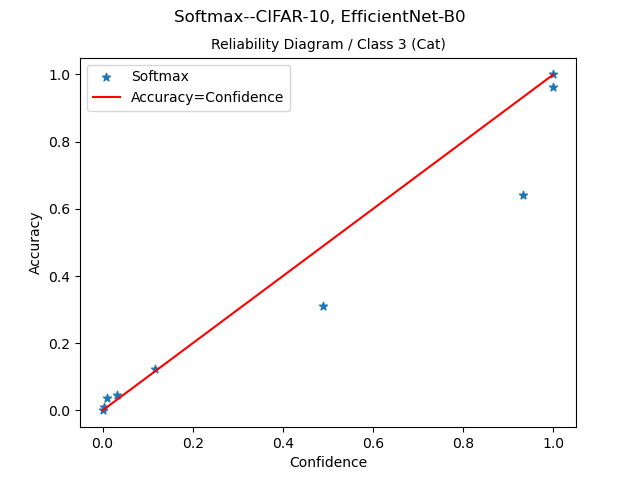}
\caption{Softmax EfficientNet-B0 ACE Reliability Diagram}
\end{subfigure}
\caption{ACE Reliability Diagrams for Softmax for ResNet-18 and EfficientNet-B0 (class 3 - cat)}
\label{fig:ACE_Reliability_Diagrams_Softmax_class3}
\end{figure}

\begin{figure}
	\centering
	\fbox{
	\includegraphics[width=0.5\textwidth]{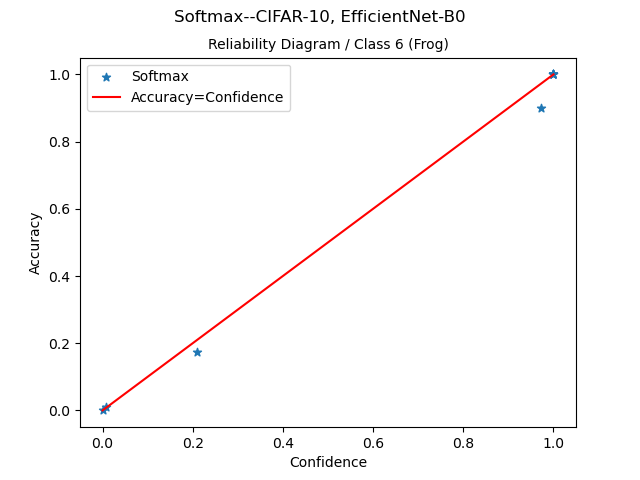}
	}
	\caption{ACE Reliability Diagram for Softmax for EfficientNet-B0 (class 6 - frog)}
	\label{fig:ACE_Reliability_Diagram_Softmax_class6}
\end{figure}

Similarly, we probe ACE edge conditions. In Figure \ref{fig:ACE_Reliability_Diagrams_Softmax_class3}(a), looking at the ACE reliability diagram for the condition where Softmax has its highest adaptive calibration error (ResNet-18/class 3 - cat). (Note this diagram uses a scatter plot vice a bar plot due to variable bin width and non-uniform bin centers, and without a histogram because bin frequencies are uniform). In this condition, Softmax demonstrates significant overconfidence over most of the range. In comparison, at the corresponding condition with a more accurate network (EfficientNet-B0/class 3 - cat) Figure \ref{fig:ACE_Reliability_Diagrams_Softmax_class3}(b) shows reduced (but still present) overprediction. Finally, at the best condition for Softmax, (EfficientNet-B0/class 6 - frog), we see in Figure \ref{fig:ACE_Reliability_Diagram_Softmax_class6} Softmax is well calibrated over the range of confidence values. These results show that reduced overprediction in Softmax corresponds with improved calibration error results.

\subsubsection{Maximum Confidence Error}

\begin{figure}[htb]
	\begin{subfigure}[h]{0.4\linewidth}
		\includegraphics[width=\linewidth]{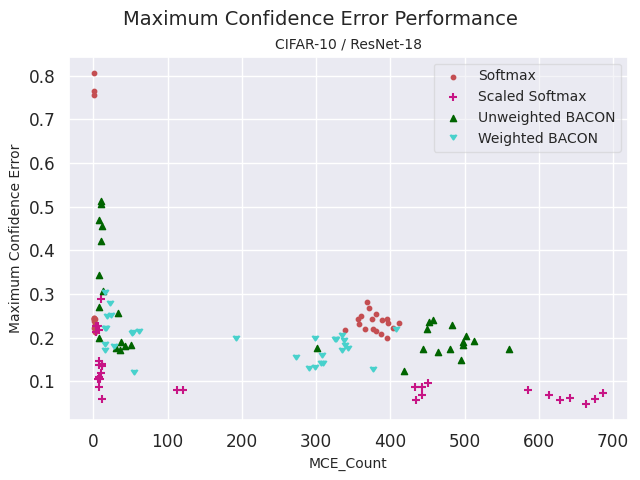}
		\caption{MCE for ResNet-18}
	\end{subfigure}
	\hfill
	\begin{subfigure}[h]{0.4\linewidth}
		\includegraphics[width=\linewidth]{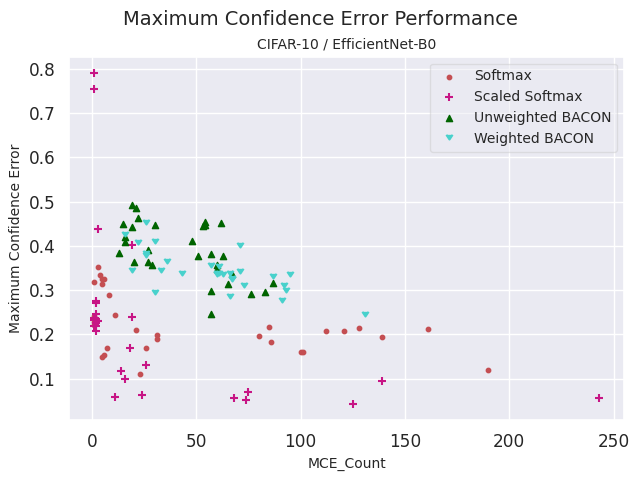}
		\caption{MCE for EfficientNet-B0}
	\end{subfigure}
	\caption{MCE results for ResNet-18 and EfficientNet-B0}
	\label{fig:MCE_Results}
\end{figure}

Results for MCE show that variability becomes a concern with higher accuracy networks. Results for ResNet-18 (accuracy $\sim$ 85\%) are shown in Figure \ref{fig:MCE_Results}{a}, while results for EfficientNet-B0 (accuracy $\sim$ 95\%) are shown in Figure \ref{fig:MCE_Results}{b}. These figures plot the MCE value against the frequency for the bin used to compute the MCE metric. For ResNet-18, almost all bins used to compute MCE have 500 or fewer counts, with a significant fraction below 100 counts. And as the bin count decreases below 100 counts, the spread of MCE values increases. The trend towards small bin counts is even more pronounced with the more accurate EfficientNet-B0. Here all bins used to compute MCE have fewer than 170 counts, with a clear trend towards increased spread in MCE values at smaller bin counts.

It is worth noting that for a uniform distribution of data across M = 9 bins, each bin would have 741 data points. In comparison, all MCE estimates originate in bins with smaller bin counts, and in many cases, much smaller bin counts, with variability increasing as bin counts drop below 100.

\subsection{Class-by-class Metrics}

\begin{figure}[htb]
	\begin{subfigure}[h]{0.4\linewidth}
		\includegraphics[width=\linewidth]{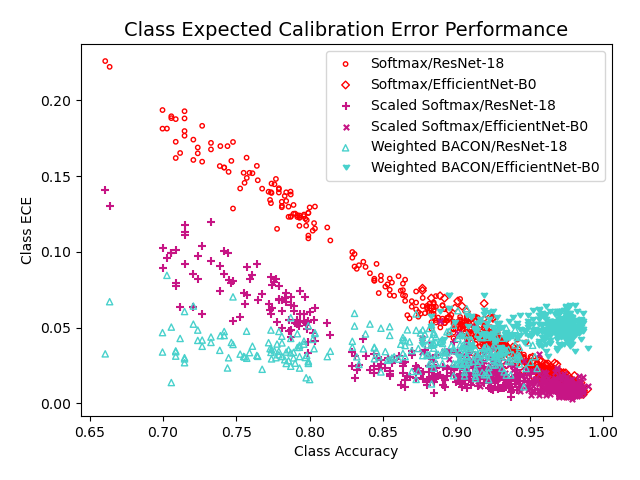}
		\caption{Class ECE vs class accuracy}
	\end{subfigure}
	\hfill
	\begin{subfigure}[h]{0.4\linewidth}
		\includegraphics[width=\linewidth]{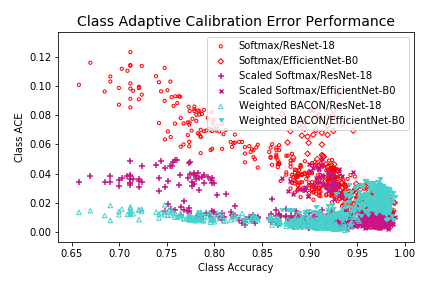}
		\caption{Class ACE vs class accuracy}
	\end{subfigure}
	\caption{Class calibration error performance as a function of class accuracy}
	\label{fig:Class_performance_vs_net_accuracy}
\end{figure}

The different results for ECE and ACE for two different networks with two different network accuracies motivates a deeper investigation into the effect of accuracy on calibration error. The different class accuracies seen in the confusion matrices in Figure \ref{fig:Confusion_Matrices} suggests investigating calibration error as a function of class accuracies. Calibration error for Softmax, T-scaled Softmax, and Weighted BACON for each class was computed with results shown in Figure \ref{fig:Class_performance_vs_net_accuracy}.

There are several interesting effects on display in Figure \ref{fig:Class_performance_vs_net_accuracy}. First, Softmax appears to have an inverse linear relationship with class accuracy in both ECE and ACE. In comparison, Weighted BACON's calibration error appears to be independent of class accuracy, with the exception of an uptick at the higher range of class accuracy. This latter effect may be due to these confidence estimates occurring at the smallest angle region of the probability density function where the rate of change is very high (see Figure \ref{fig:Output_Angle_Distribution}). T-scaled Softmax also has an inverse dependence on class accuracy, outperforming Weighted BACON at higher accuracies, but performing worse at lower accuracies. This class accuracy effect appears to explain the different comparative results between confidence estimators for the cases of ResNet-18 and EfficientNet-B0. For ResNet-18, data points tend to reside at lower accuracy where Weighted BACON outperforms Softmax and T-Scaled Softmax, while EfficientNet-B0 data points tend to reside at higher accuracy, where Softmax is competitive with, and T-scaled Softmax generally outperforms Weighted BACON. Another interesting effect present in Figure \ref{fig:Class_performance_vs_net_accuracy} is that for Softmax, the ECE results appear to have less deviation from the trendline than for ACE. In fact, for the case of ACE, it looks like Softmax results for ResNet-18 and EfficientNet-B0 may have different trendlines. In comparison for BACON, ACE results appear to have less deviation from the trendline than for ECE, and in both cases, ResNet-18 and EfficientNet-B0 results appear similar in their overlap region. These features, not obvious when summed over all classes, provide useful insight for characterizing confidence estimator performance.

\section{Discussion}

\textit{Metrics Characterization}  

We used three metrics, ECE, ACE, and MCE to characterize the confidence estimator peformance for two different neural networks on our imbalanced CIFAR-10 dataset. ECE and ACE showed different results for the same conditions, while MCE results were examined closely. We discuss the performance of each metric in turn.

ECE provided different results for ResNet-18 or EfficientNet-B0. For both cases, T-scaled Softmax was the superior confidence estimator. However, for the the less accurate ResNet-18, weighted and unweighted BACON outperformed unscaled Softmax, while for the more accurate EfficientNet-B0, unscaled Softmax outperformed weighted and unweighted BACON. A possible explanation is that this is a result of being a fixed bin width metric. Softmax is well known to yield overpredictions. As network accuracies increase, these overpredictions are increasingly shifted to the rightmost bin. When this occurs, ECE is no longer able to resolve overpredictions within the bin. In addition, the shifting of overpredictions to the right depopulates other bins with overpredictions, reducing the population weighting associated with the accuracy/confidence gaps in the ECE calculation. This effect should be expected to be more pronounced at higher classification accuracies, and we do see Softmax performance improve between the 85\% accuracy ResNet-18 results and the 95\% EfficientNet-B0 results. Also worth noting is that ECE did not resolve the expected performance difference from Bayesian weighting of the BACON confidence estimates.

As an adaptive binning metric, ACE showed different, and interesting results. Here, both weighted and unweighted BACON outperformed unscaled Softmax for both ResNet-18 and EfficientNet-B0. Comparison between between BACON and T-scaled Softmax showed that BACON was slightly superior at lower accuracy (ResNet-18) but slightly worse at higher accuracy (EfficientNet-B0). In addition, there was a slight upward in trend in ACE with classification accuracy for BACON. However, the key result here is that the ACE metric resolves the expected behavior of improving confidence estimation performance by using Bayesian weights in the BACON estimator.

Further understanding of these results are given by considering the individual class calibration error performance results. For both ECE and ACE, unscaled and T-scaled Softmax show a downward trend in calibration error as class accuracy increases, while weighted BACON is relatively constant across the entire range of operating conditions. The change of performance for ECE and ACE over the entire test dataset as we move from the less accurate ResNet-18 to the more accurate EfficientNet-B0 reflect an averaging of results over different operating conditions. ResNet-18 results reflect averaging over operating conditions more favorable for BACON, while EfficientNet-B0 results reflect averaging over operating conditions more favorable to T-scaled Softmax.

This dependence on operating conditions is an important result. Calibration robustness over a wide range of operating conditions is desirable characteristic of a confidence estimator. Here we've seen that BACON, based on the geometric representation of the neural network's terminal layers is robust in these experimental conditions, while Softmax and T-scaled Softmax calibration appears to be a function of network accuracy. The Softmax estimators appear to perform best at the highest network accuracies, but that is where a confidence estimator is needed least. The important region for confidence estimation is at lower classification accuracies, and here BACON performs best.

Another important result is the tendency for MCE to be determined from low count fixed width bins as neural networks increase in accuracy. In addition, as bin count decreases, MCE variability increases. This tendency to compute a noisy metric from a small sample size bin brings into question the utility of this metric as neural networks continue to increase in performance.

\textit{Confidence Estimator Performance}

When comparing calibration error between Softmax and BACON, regardless of using the ECE or ACE metric, the salient feature is the inverse linear dependence of Softmax on classification accuracy, while BACON is relatively insensitive to classification accuracy (see Figure \ref{fig:Class_performance_vs_net_accuracy}(a,b). As shown in Figures \ref{fig:ACE_Reliability_Diagrams_Softmax_class3}(a,b) and \ref{fig:ACE_Reliability_Diagram_Softmax_class6}, this reduction in calibration error with increased network accuracy results from a decrease in overprediction, which is itself results in compression of predicted confidences to higher levels of confidence, leaving less room for overprediction to occur.

This implies Softmax's confidence error has a systematic component, while BACON's confidence error is mostly stochastic in nature. The origin of the systematic confidence error in Softmax is likely rooted in the origin of the Softmax function. It is asserted to be a probability distribution function, yet was not derived as a probability distribution function.  Bridle \cite{Bridle:1989} introduced Softmax to constrain network output values to provide a conditional probability distribution. But this does so in the aggregate, by minimizing loss over a set of data. Thus it should be expected that calibration error should in aggregate have behavior that reflects overall network performance, because Softmax does not estimate probability based on local information. And its connection to the Boltzmann Probability Distribution e.g.  \cite{Tokic:2011,Tijsma:2016} was misplaced, as the problem at hand does not satisfy the requirements for equilibrium. A simple examination of output node angle distributions in Figure \ref{fig:Output_Angle_Distribution} reveals the necessary condition of equal probability of microstates, required for Boltzmann equilibrium \cite{Reif:1965}, is not satisfied. 

In comparison, BACON is derived from Bayes' Rule, and its error should be mostly associated with how well the angle distributions derived from the validation data set correspond to the test set, as well as the amount of evidence available to support the estimate. And we see in Figure \ref{fig:Class_performance_vs_net_accuracy} that BACON calibration performance is relatively flat, except in the very highest accuracy regime, which corresponds with approaching the lower angle limit of the rapidly changing output angle distribution (see Figure \ref{fig:Output_Angle_Distribution}). Yet beyond this edge case of very high accuracy, BACON appears to outperform Softmax in calibration error, whether measured using ECE or ACE metrics, over the lower accuracy operating conditions where confidence estimation is most valued.  

\section{Conclusion}

This paper introduces the Bayesian Confidence Estimator (BACON), which estimates the prediction confidence for each output node. BACON estimates confidences by applying Bayes' Rule to the geometric representation of the terminal layers of a deep neural network. This differs from current practice using Softmax, which uses an exponential transform to constrain outputs between 0 and 1, and ensure all output values sum to 1. 

We compared performance of the BACON algorithm to both unscaled and temperature-scaled Softmax using the CIFAR-10 dataset, with an imbalanced test dataset, in a variety of operating conditions. These included two distinct CNN architectures, as well as thirty different train/val splits in order to generate z-distribution statistics and report confidence intervals.   

Analysis showed BACON provided stable and low calibration error over the range of operating conditions. BACON outperformed unscaled Softmax with both ECE and ACE metrics over most operating conditions, with the exception of the edge case of very high classification accuracy. Softmax calibration errors were inversely linearly proportional to classification accuracy, while BACON calibration errors were insensitive to classification accuracy, except at very high classification accuracy where calibration error worsened slightly. These results were mostly consistent whether using the ECE or ACE metric. In comparison, temperature-scaled Softmax outperformed BACON at high accuracies, but performance degraded compared to BACON at lower classification accuracies.

BACON also demonstrated a novel capability using weighted inputs to improve calibration error performance on an imbalanced test set. This weighted BACON estimator outperformed the uniformly weighted BACON estimator, when using the ACE metric. We note that our results are optimistic, as we used actual class weights in our calculation, which will not be available in a real-world setting. However, demonstrating expected improvement in calibration error when using actual weights is an important test result at a known boundary condition.

It was also shown that MCE is increasingly variable and untrustworthy under conditions of high classification accuracy (>85\%) due to small sample size effects. Under these conditions, MCE was calculated using bins with lower frequencies, resulting in increased variability as bin frequencies decreased into the 10s of counts.

Finally, the validity and utility of the geometric representation introduced in \cite{Waagen:2020} was further supported in this experiment, by using its coordinate system in calculating Bayesian probabilities for neural network outputs, resulting in calibration errors that were robust over most operating conditions in this experiment. 

This paper has several strengths and limitations, both of which present opportunities for follow-on research. This paper is the result of experiments on two CNN architectures, each with 30 randomly chosen seeds, which enabled estimation and reporting of calibration error confidence intervals. This diversity of experimental conditions also enabled the discovery of small sample size effects on MCE estimation. However, this paper only discusses results on a single dataset. This paper does not discuss small or imbalanced training sets, classes that are not as easily separable, or datasets which contain a much larger number of classes (e.g., CIFAR-100). The latter is expected to pose greater compute requirements. Finally, our use of actual class distribution fractions as Bayesian weights, while useful in demonstrating the concept, also provides the most optimistic case for performance improvement. Sensitivity analysis is yet to be performed.  

Multiple paths for future research lie ahead. These include: 1) Experimentation on more challenging datasets; 2) Application of other Bayesian frameworks to the geometric representation, with a focus on improving performance in the high classification accuracy region; 3) Exploration into potential robustness in the presence of data drift; and 4) Exploration of the origin of systematic errors in Softmax. We are working to make progress in these areas.

\section{Acknowledgements}

This work was financially supported by the Air Force Research Laboratory with funding supporting student intern researchers at the Advanced Technology Research Center at the University of Dayton and Wright State University (contracts FA8650-23-2-1089 and FA8650-18-2-1645). This work was also supported in part by high-performance computer time and resources from the DoD High Performance Computing Modernization Program. The authors also wish to thank W.F. Bailey, C.B. Liberatore, C.J. Menart, O.L. Mendoza-Schrock, K.L. Priddy, V.J. Velten, D.E. Waagen, and E.G. Zelnio for helpful insights.

\bibliographystyle{ieeetr}
\bibliography{refs}  

\end{document}